\documentclass[10pt,twocolumn,letterpaper]{article}

\usepackage{iccv}
\usepackage{times}
\usepackage{epsfig}
\usepackage{graphicx}
\usepackage{amsmath}
\usepackage{amssymb}
\usepackage{color}
\usepackage[dvipsnames]{xcolor}
\usepackage{arydshln} % cdashline
\usepackage{hhline}
\usepackage[font=small]{caption}
\usepackage{subcaption}
\usepackage{booktabs, multirow}
\usepackage{pifont}% http://ctan.org/pkg/pifont
\usepackage{enumitem}
\newcommand{\cmark}{\ding{51}}%
\definecolor{citecolor}{HTML}{0071bc}

\newcommand{\unitablestyle}{\tablestyle{6.5pt}{1.1}}%
\usepackage{color}
\usepackage{xcolor}
\definecolor{citecolor}{HTML}{0071bc}

\usepackage[pagebackref=true,breaklinks=true,letterpaper=true,colorlinks,citecolor=citecolor,bookmarks=false]{hyperref}

\iccvfinalcopy % *** Uncomment this line for the final submission

\def\iccvPaperID{}% *** Enter the ICCV Paper ID here

\newcommand*{\rowstyle}[1]{% sets the style of the next row
 \gdef\@rowstyle{#1}%
 \@rowstyle\ignorespaces%
}

\newlength\savewidth\newcommand\shline{\noalign{\global\savewidth\arrayrulewidth
  \global\arrayrulewidth 1pt}\hline\noalign{\global\arrayrulewidth\savewidth}}
\newcommand{\tablestyle}[2]{\setlength{\tabcolsep}{#1}\renewcommand{\arraystretch}{#2}\centering\footnotesize}
\makeatletter\renewcommand\paragraph{\@startsection{paragraph}{4}{\z@}
  {.5em \@plus1ex \@minus.2ex}{-.5em}{\normalfont\normalsize\bfseries}}\makeatother

\usepackage{algorithm}
\usepackage{listings}

\usepackage{etoolbox}
\makeatletter
\AfterEndEnvironment{algorithm}{\let\@algcomment\relax}
\AtEndEnvironment{algorithm}{\kern2pt\hrule\relax\vskip3pt\@algcomment}
\let\@algcomment\relax
\newcommand\algcomment[1]{\def\@algcomment{\footnotesize#1}}
\renewcommand\fs@ruled{\def\@fs@cfont{\bfseries}\let\@fs@capt\floatc@ruled
  \def\@fs@pre{\hrule height.8pt depth0pt \kern2pt}%
  \def\@fs@post{}%
  \def\@fs@mid{\kern2pt\hrule\kern2pt}%
  \let\@fs@iftopcapt\iftrue}
\makeatother

\ificcvfinal\fi

\begin{document}

%%%%%%%%% TITLE
\title{Contrastive Learning of Image Representations with \\ Cross-Video Cycle-Consistency}

\author{Haiping Wu\\
McGill University, Mila
% {\tt\small haiping.wu2@mail.mcgill.ca}
\and
Xiaolong Wang\\
UC San Diego
% {\tt\small xiw012@ucsd.edu}
}

\maketitle
\ificcvfinal\thispagestyle{empty}\fi

\begin{abstract}
    Recent works have advanced the performance of self-supervised representation learning by a large margin. The core among these methods is intra-image invariance learning.
    Two different transformations of one image instance are considered as a positive sample pair, where various tasks are designed to learn invariant representations by comparing the pair. Analogically, for video data, representations of frames from the same video are trained to be closer than frames from other videos, i.e. intra-video invariance. However, cross-video relation has barely been explored for visual representation learning. Unlike intra-video invariance, ground-truth labels of 
    cross-video relation is usually unavailable without human labors. In this paper, we propose a novel contrastive learning method which explores the cross-video relation by using cycle-consistency for general image representation learning. This allows to collect positive sample pairs across different video instances, which we hypothesize will lead to higher-level semantics. We validate our method by transferring our image representation to multiple downstream tasks including visual object tracking, image classification, and action recognition. We show significant improvement over state-of-the-art contrastive learning methods. Project page is available at \url{https://happywu.github.io/cycle_contrast_video} .
    
\end{abstract}

%%%%%%%%% BODY TEXT
\section{Introduction}

There has been a surge of recent interest in contrastive learning of visual representation~\cite{wu2018unsupervised,hjelm2018learning, bachman2019learning,henaff2019data,tian2019contrastive, chen2020simple,he2020momentum,misra2020self}. We have witnessed that contrastive learning out-performs supervised pre-training with large-scale human annotations in various visual recognition tasks~\cite{he2020momentum,chen2020simple}. The key of this self-supervised task is to construct different views and transformations of the same instance, and learn the deep representation to be invariant to the view changes. To construct different views for forming positive image pairs in contrastive learning, the most common way is to use different data augmentations on the same instance (\eg random cropping, image rotation, colorization). 

  \begin{figure}[t]
        \centering
        \includegraphics[width=0.5\textwidth]{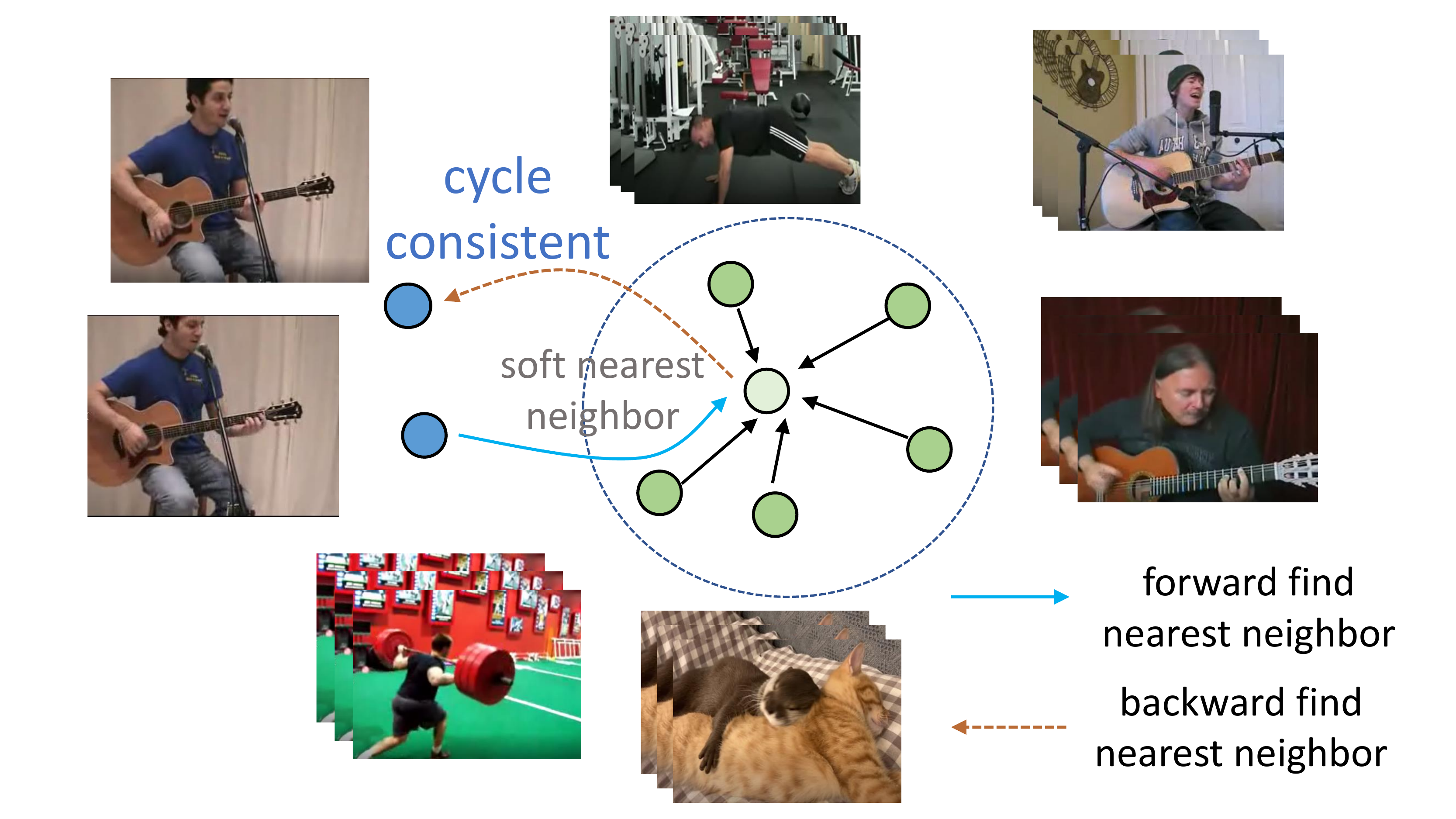}
        \caption{Cross-video cycle-consistency for image representation learning. Starting from one frame in a video, we find its soft nearest neighbor from other videos as a forward step, then the cycle-consistency is achieved when the soft nearest neighbor finds its closest frame within the same video as the start frame in a backward step.
        }
        \label{fig:cycle}
    \end{figure}

However, simply performing artificial augmentation on single instance has shown its limitation in multiple applications~\cite{tian2020makes,xiao2020should}. For example, Tian \etal~\cite{tian2020makes} have performed detailed analysis on how different augmentations can affect different downstream visual recognition tasks. Going beyond single image, researchers have also looked into videos as the source for obtaining positive pairs of training images~\cite{purushwalkam2020demystifying,gordon2020watching,Wang_UnsupICCV2015}. That is, two nearby frames in the same video can be taken as a natural augmentation in time for the same object instance. By training using temporal augmentation, the representation can learn viewpoint and deformation invariance. However, these approaches are still limited to find positive pairs and learning their similarity within a single instance. 

In this paper, we propose to perform contrastive learning with positive image pairs sampled across different videos instead of the same video. We hypothesize this can potentially capture higher-level semantics and categorical information beyond low-level intra-instance invariances modeled by previous approaches. Specifically, given two image frames $I_i$ and $I_j$ from a video, instead of directly using them as a positive pair for training~\cite{purushwalkam2020demystifying,gordon2020watching}, we will first ``explain'' frame $I_i$ by composing frames from other videos that are similar to $I_i$, then compare the composed frames to $I_j$ for contrastive learning.

Assuming we have a neural network feature extractor to learn, we extract the feature representations for the image frame $I_i$ as $q_i$, and representations for frames from other videos as $U=\{u_1, u_2, ..., u_m\}$. Given these representations, we compute the similarities between $q_i$ and $U$, and normalize them to a probability distribution. We use this probability distribution to re-weight and compose the features $U$ as a new feature representation for frame $I_i$ (frames that are more similar to $I_i$ will have larger weight). We call this new feature as a soft nearest neighbor to $I_i$. We then form a positive pair of training data with this new representation and the feature of $I_j$ (a different frame from the same video as $I_i$). As shown in Figure~\ref{fig:cycle}, this procedure goes through a cycle of starting from one frame $I_i$ in a video, searching forward by matching frames from other videos, and retrieving frame $I_j$ backward in the first video. We call this process \emph{Cycle-Consistent Contrastive Learning}. Intuitively, enforcing such a cycle-consistency can explicitly push video frames with similar structure closer, thus leads to a natural clustering of semantics.

We perform the proposed self-supervised representation learning on unlabeled video dataset Random Related Video Views (R2V2)~\cite{gordon2020watching} and transfer the learned representation to various downstream tasks including visual object tracking, image classification and action recognition. We stress our goal is to use the temporal signal to learn a \textbf{general image-level representation} for multiple applications beyond videos-level recognition tasks. We show significant improvements over multiple state-of-the-art approaches. We also conduct extensive ablation studies of different components and design choices of our method.

Our contributions include: (i) A novel cross-video cycle-consistent contrastive learning objective that explores cross-video relations, going beyond previous intra-image and intra-video invariant learning; 
(ii) The proposed loss enforces image representations from the same category (of similar visual structures) closer without explicitly generating pseudo labels; (iii) The learned image representation achieves significant improvement in multiple downstream tasks including object tracking, image classification and action recognition.

\section{Related Work}

\textbf{Contrastive learning.}
The self-supervised contrastive learning methods~\cite{hadsell2006dimensionality, dosovitskiy2014discriminative,wu2018unsupervised,hjelm2018learning, oord2018cpc, bachman2019learning,henaff2019data,tian2019contrastive,zhuang2019local, chen2020simple,he2020momentum,misra2020self, chen2020mocov2,caron2020swav} try to learn image representations under different transformations agree by forming positive and negative pairs, and make the representations of positive pairs have high similarity and negative pairs have low similarity. The typical way of generating positive pairs is performing artificial data augmentations on a single image instance. For example, Chen \etal~\cite{chen2020simple} introduce a contrastive learning baseline with different types of augmentations, including random cropping, resizing, color distortion, gaussian blur, \etc. He \etal~\cite{he2020momentum} proposes MoCo which introduces a momentum network to encode a queue of a large number of egative samples for efficient learning. In this work, we build our model based on the  MoCo framework. However, instead of learning with positive pairs by augmenting the same image, we propose a new objective which finds positive pairs of sample across videos for contrastive learning of image representations.

\textbf{Self-supervised image representation learning from videos.} Going beyond learning from a single image~\cite{dosovitskiy2015discriminative,doersch2015unsupervised,pathak2016context,donahue2016adversarial,zhang2016colorful,gidaris2018unsupervised}, video naturally offers temporal information and multi viewpoints for objects, which have been extensively utilized as self-supervisory signals in representation learning~\cite{Goroshin2015,Agrawal2015,Jayaraman2015,Wang_UnsupICCV2015,pathak2017learning,owens2016ambient,misra2016shuffle,wang2017transitive,lee2017unsupervised,buchler2018improving,wei2018learning,jing2018self,sayed2018cross}. For example, Wang and Gupta~\cite{Wang_UnsupICCV2015} use tracking to provide supervision signals that makes the feature representations of tracked patches similar. Recent works~\cite{gordon2020watching,purushwalkam2020demystifying,jabri2020walk,xu2021rethinking} further extend similarity learning between video frames under the contrastive learning framework, where the positive pairs for training are frames sampled from the same video. It has been shown image representations with viewpoint invariance can be learned. Our work is motivated by these previous works, and going beyond viewpoint invariance, learning using positive pairs across videos can potentially lead to image representations with higher-level semantics. While contrastive learning has also been applied to video representation learning with 3D ConvNets for action recognition~\cite{jing2018self,han2019video,kim2019self,benaim2020speednet,qian2020spatiotemporal,wang2020self,han2020memory,jenni2020video,morgado_avid_cma}, we emphasize our work is focusing on learning a general image representation for multiple tasks beyond action recognition including visual tracking and image classification.

\textbf{Cycle-consistency learning.} Our work is influenced by cycle-consistency learning in different computer vision applications including 3D scene understanding~\cite{huang2013consistent,zhou2017unsupervised,godard2017unsupervised,yin2018geonet}, image alignment and translation~\cite{zhou2015flowweb,zhou2015multi,zhou2016learning,zhu2017unpaired}, and space-time alignment in videos~\cite{Recycle-GAN,wang2019learning,lai2019self,dwibedi2019temporal,wang2019udt,jabri2020walk,purushwalkam2020aligning,kong2020cycle}. For example, Wang \etal~\cite{wang2019learning} propose to perform forward and backward tracking in time to achieve a cycle-consistency for learning temporal correspondence.  Dwibedi \etal~\cite{dwibedi2019temporal} formulates a temporal cycle consistency loss which aligns frames from one video to another between a pair of videos, and achieves good performance in video frame alignemnt tasks. Building on these two works, Purushwalkam \etal~\cite{purushwalkam2020aligning} propose to track object patches inside a video and align them across videos at the same time. While these results are encouraging, both approaches learning from video pairs~\cite{dwibedi2019temporal,purushwalkam2020aligning} require human annotators to provide ground-truth pairs (video-level) in training with a small scale of videos. In this paper, we propose to go beyond these restrictions, and apply cross-video cycle-consistency learning without any human annotations. These not only allows learning with a large-scale videos, but also generalizes our representations to multiple downstream vision tasks.

\section{Cycle-Consistent Contrastive Learning}

In this section, we first introduce contrastive learning with different forms of invariant learning targets. Then we propose our method with cross-video cycle-consistency learning.

\subsection{Intra-Image and Intra-Video Invariance}

The core of self-supervised contrastive learning~\cite{he2020momentum,chen2020simple} is to learn representations which maximize the agreement between different views, augmentations of one image  instance, and minimize the similarity between two different and unrelated instances at the same time. Most methods share the similar learning objective, which is to make the representations \emph{intra-image} invariant. We describe the formulation of the objective as follows.

Given an query image $I_{i}$, the feature extractor encodes it to feature representations $q_i$ and $k_i$ under two different data augmentations. The intra-image invariance learning considers $q_i$ and $k_i$ as 
an a positive training pair and minimizes their representation distance, while maximizing the representation distance of $q_i$ and a set of negatives $U = \{u_1, u_2, ..., u_m\}$, which is a set of feature representations extracted from different images. The intra-image invariance contrastive learning loss function is defined as,
\begin{equation}
    \mathcal{L}_{\text{intra-image}} =-\log \frac{\exp \left(\operatorname{sim}\left(q_i, k_i \right) / \tau\right)}{\sum_{u \in \{U, k_i\}} \exp \left(\operatorname{sim}\left(q_i, u\right) / \tau\right)},
    \label{eq:intra-image}
\end{equation}
where $\tau$ is the temperature constant and 
$\operatorname{sim}(x, y)=x^{\top} y /\|x\|\|y\|$ is the cosine similarity between two feature vectors. The loss function tries to classify $q_i$ is similar to $k_i$ from the same image against features in $U$ from different images, achieving the intra-image invariance learning.

  \begin{figure}[t]
        \centering
        \includegraphics[width=0.50\textwidth]{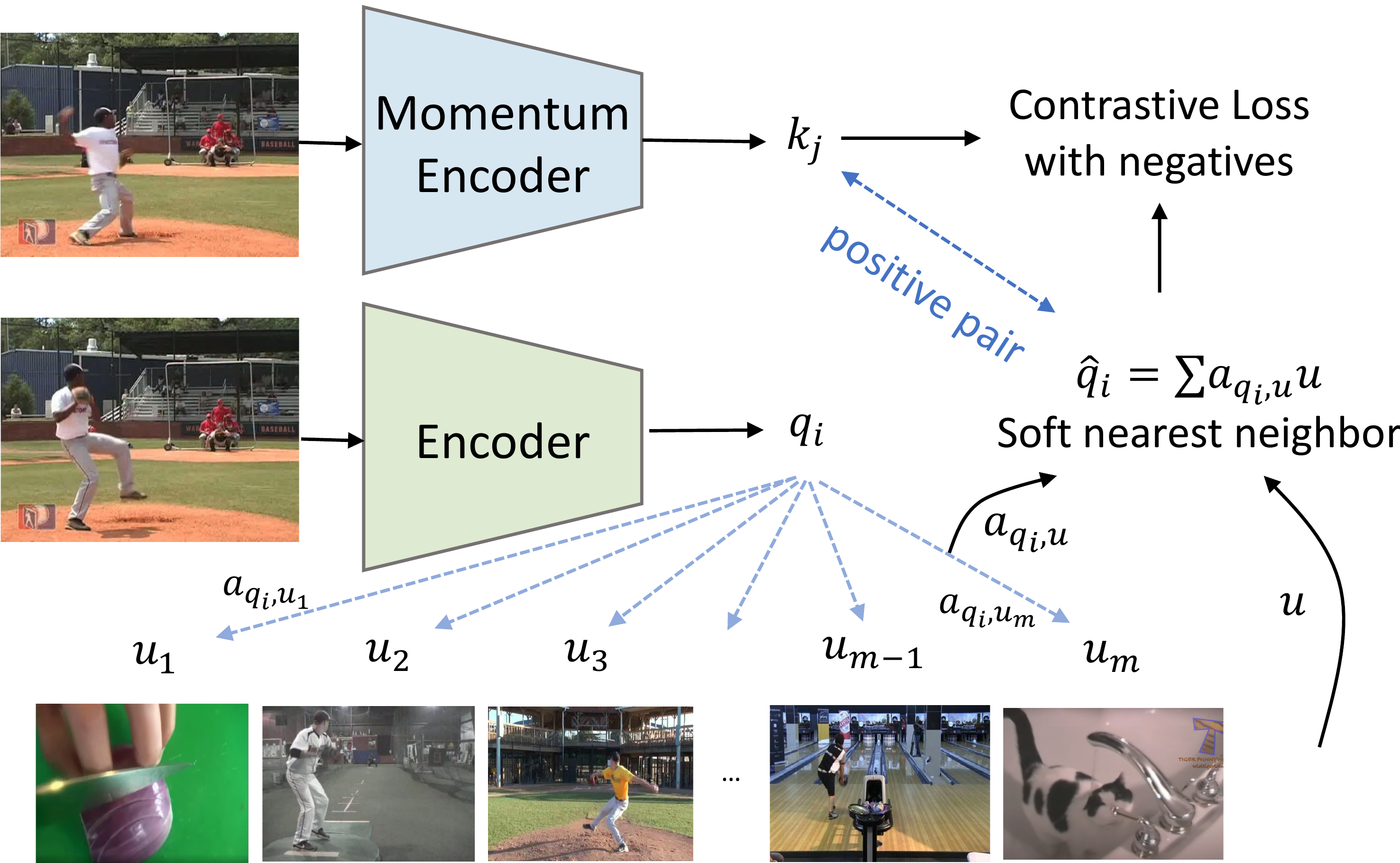}
        \vspace{-0.1in}
        \caption{
            The pipeline for the proposed cross-video cycle-consistency loss $\mathcal{L}_{\text{cycle}}$. $U=\{u_1, u_2, ..., u_m\}$ is the neighbor representation set. Given the query feature $q_i$, we construct its soft nearest neighbor $\hat{q_i}$ by combining frame information from other videos. We use the key feature $k_j$ from the same video and $\hat{q_i}$ as a positive pair for contrastive learning.
        }
        \label{fig:pipeline}
    \end{figure}

While intra-image invariance learning gives us good representations, augmentations of static images fail to capture viewpoint and deformation variations of instances (\eg different viewpoints or gestures of one person)~\cite{purushwalkam2020demystifying}. It is natural to resort to sequential video data that has variations of the same instance across time, extending the \emph{intra-image} invariance learning to \emph{intra-video} invariance learning. The intra-video invariance learning
considers frames within the same video in a local time window as a invariant set to form positive pairs in training~\cite{sermanet2018time,gordon2020watching,purushwalkam2020demystifying}. Similarly to Eq.~\ref{eq:intra-image}, the intra-video invariance contrastive learning loss is defined as,
\begin{equation}
    \mathcal{L}_{\text{intra-video}} =-\log \frac{\exp \left(\operatorname{sim}\left(q_i, k_j \right) / \tau\right)}{\sum_{u \in \{U, k_j\}} \exp \left(\operatorname{sim}\left(q_i, u\right) / \tau\right)},
    \label{eq:intra-video}
\end{equation}
where $k_j$ is the feature representation of image $I_j$, which is sampled from the same video of $I_i$. $\mathcal{L}_{\text{intra-video}}$ tries to make the feature representations of images from the same video 
closer than images from different videos. When the sampled image $I_j$ is the same as $I_i$, $\mathcal{L}_{\text{intra-video}}$ is identical to $\mathcal{L}_{\text{intra-image}}$. We can see the intra-image invariance learning as a special case of the intra-video invariance learning.

\subsection{Cross-Video Cycle-Consistency Objective}
Both intra-image and intra-video invariance learning make the feature representations of the same instance (from the same image or video) closer. However, there is no explicit regularization on the distances between representations from the same class (or images of similar visual structures). For instance, the representations of different cat instances should be close 
and representations of frames from different videos on playing tennis should also be similar. In this section, we propose to find cross-video correspondence with  cycle-consistency without using any ground-truth labels, and incorporate the correspondence in contrastive learning. The pipeline of the new proposed objective is shown in Figure~\ref{fig:pipeline} and we will introduce the formulations as following.

Our new objective consists of a forward and a backward nearest neighbor processes. Given a encoded query feature representation $q_i$ of $I_i$ from a video $V$, we first compute its nearest neighbor $\hat{q}_i$ in a candidate set $U$ containing frames from any videos. Then, we find the nearest neighbor of $\hat{q}_i$ backwards within the union set of $U$ and $V$. We emphasize that $U$ \emph{does not} include any frames from the video $V$. The cycle-consistency is achieved when the backward nearest neighbor of $\hat{q}_i$ is in the desired invariance learning set of $q_i$ from $V$. In order to make the learning differentiable, we propose to compute the soft nearest neighbor as the \emph{forward step} in our objective. Formally,
the soft nearest neighbor $\hat{q}_i$ of $q_i$ in $U = \{u_1, u_2, ..., u_m\}$ is calculated as,
\begin{equation}
    \hat{q}_i = \sum_{u\in U} \alpha_{q_i, u} u,
    \label{eq:cycle_forward}
\end{equation}
where $\alpha_{q_i, u}$ is the normalized similarity of $q_i$ and $u$, which is defined as 
\begin{equation}
    \alpha_{q_i, u} = \frac{\exp(\operatorname{sim}(q_i, u)/\tau)}{\sum_{u' \in U} \exp (\operatorname{sim}(q_i, u')/\tau)},
    \label{eq:cycle_sim}
\end{equation}
where $\tau$ is the temperature and 
$\operatorname{sim}(x, y)=x^{\top} y /\|x\|\|y\|$
is the cosine similarity. 

Given the soft nearest neighbor $\hat{q_i}$, we assume if a representation is good for understanding the high-level semantics,  $\hat{q_i}$ should be in the invariance set of $q_i$. Recall that we denote the invariance target of $q_i$ as $k_j$ (feature of a different frame in the same video) in intra-video invariance learning. We perform non-parametric 
classification as the \emph{backward step}, and the cross-video cycle contrastive loss is defined as,
\begin{equation}
    \mathcal{L}_{\text{cycle}} =-\log \frac{\exp \left(\operatorname{sim}\left(\hat{q}_i, k_j \right) / \tau\right)}{\sum_{u \in \{U, k_j\}} \exp \left(\operatorname{sim}\left(\hat{q}_i, u\right) / \tau\right)},
    \label{eq:cycle_back}
\end{equation}
where $k_j$ is the feature representation of image $I_j$ from the invariant set of $I_i$, and we consider $I_j$ and $I_i$ are sampled from the same video $V$.

Intuitively, Eq.~\ref{eq:cycle_forward} tries to use the representations in the candidate set to reconstruct the query feature representation, according to the similarity measurement of Eq.~\ref{eq:cycle_sim}. Then $\mathcal{L}_{\text{cycle}}$ 
is minimized when the reconstructed feature representation is close to the representations of another image from video $V$. 
In this way, the model will learn to find correspondence across videos where images with similar visual structures are encouraged to be the same, and invariant within a video at the same time. By building the correspondence across videos, it can potentially help the representations to learn category-level information.

The overall learning target is a combination of intra-video invariant loss $\mathcal{L}_{\text{intra-video}}$ and cross-video cycle consistency loss $\mathcal{L}_{\text{cycle}}$, defined as 
\begin{equation}
    \mathcal{L} = \mathcal{L}_{\text{intra-video}} + \lambda\mathcal{L}_{\text{cycle}},
    \label{eq:loss}
\end{equation}
where $\lambda$ is a balancing factor of the two learning targets. We will provide ablation on $\lambda$ in our experiments.

\paragraph{Implementation details for neighbor set $U$.} 

In our experiments, we use two separate nearest neighbor sets for \emph{forward and backward steps} for our objective $\mathcal{L}_{\text{cycle}}$. The nearest neighbor set $U$ in Eq.~\ref{eq:cycle_forward} is selected by randomly sampling from the current memory bank \cite{he2020momentum} at each training iteration. The remaining elements in the memory bank are used as negative candidates for finding the nearest neighbor backwards in Eq.~\ref{eq:cycle_back}. We provide the pseudo code of the $\mathcal{L}_{\text{cycle}}$ in Alg.~\ref{alg:code}.

\begin{algorithm}[t]
\caption{Pseudocode for $\mathcal{L}_{\text{cycle}}$, PyTorch-like}
\label{alg:code}
\definecolor{codeblue}{rgb}{0.25,0.5,0.5}
\definecolor{codekw}{rgb}{0.85, 0.18, 0.50}
\algcomment{\fontsize{7.2pt}{0em}\selectfont \texttt{matmul}: matrix multiplication;  \texttt{cat}: concatenation.
%\vspace{-1.em}
}
\lstset{
  backgroundcolor=\color{white},
  basicstyle=\fontsize{7.2pt}{7.2pt}\ttfamily\selectfont,
  columns=fullflexible,
  breaklines=true,
  captionpos=b,
  commentstyle=\fontsize{7.2pt}{7.2pt}\color{codeblue},
  keywordstyle=\fontsize{7.2pt}{7.2pt},
%  frame=tb,
}

\begin{lstlisting}[language=python]
# f_q: encoder + projection mlp
# f_k: momentum encoder + projection mlp
# mon: momentum parameter
# queue: key dictionary (c x K)
# m: size of neighbor set 
# t: temperature 

f_k.params = f_q.params

for x_i, x_j in loader:  # load a minibatch with n samples, each sample contains two randomly selected frames from one video
    x_i, x_j = aug(x_i), aug(x_j)  # random augmentation
    q_i, k_j = f_q(x_i), f_k(x_j)  # feature extraction, n x c
    k_j = k_j.detach()
    
    u = random.choice(queue, m) # randomly select neighbor set
    r_queue = queue \ u # get remain elements of queue
    a = softmax(cos_similarity(q_i, u)/t)  # calculate similarity: n x m
    soft_nn = matmul(a, u)  # get soft nearest neighbor
    
    logits = matmul(soft_nn, cat([k_j.T, r_queue], dim=1)) 
    labels = zeros(n)
    cycle_loss = CrossEntropyLoss(logits/t, labels)  # calculate cycle back classification loss
    
    enqueue(queue, k_j)
    dequeue(queue)
    
\end{lstlisting}
\end{algorithm}

\section{Experiments}
In this section, we conduct experiments to perform unsupervised representation learning using the proposed learning objectives. We show that 
the learned representation transfer well to various downstream tasks. Then we design extensive ablation experiments to study the effectiveness of 
the proposed cross-video cycle-consistent learning method. 

\subsection{Experiments settings}

\paragraph{Dataset}. We perform unsupervised representation learning on Random Related Video Views (R2V2)~\cite{gordon2020watching} dataset, which is a large-scale diverse video frames collection. It has 2.7M videos, with 4 frames for each video. For smaller models and ablation studies, we use a subset of it for time efficient, which has 109k videos and 438k frames in total, which we refer as R2V2-S. 

\paragraph{Network Architecture}. ResNet-18~\cite{he2016deep} is used as backbone for major ablation studies for its efficiency and accuracy. One fully connected layer (512 x 64) is used as projection layer after the global average pooling layer to 
obtain the embedding features. ResNet-50 is also adapted for comparing with other methods. 
Following the study in~\cite{xiao2020should}, $\mathcal{L}_{\text{intra-video}}$ and $\mathcal{L}_{\text{cycle}}$ use separate projection layers as they have different invariance learning targets. Following MoCo~\cite{he2020momentum}, we use 
a query encoder network $f_q$ and a key encoder network $f_k$, where the parameters $\theta_k$ of $f_k$ is updated by the parameters $\theta_q$ of $f_q$, using $\theta_{\mathrm{k}} \leftarrow m \theta_{\mathrm{k}}+(1-m) \theta_{\mathrm{q}}$.
The momentum coefficient $m$ is set to 0.999 and the memory bank size for classification is 65536. The temperature $\tau$ is set to 0.07.

\paragraph{Training}. We use SGD to optimize the unsupervised representation learning for a total of 200 epochs. The mini-batch size is 256 in 8 GPUs. The initial learning rate is 0.06. Other training recipes follow~\cite{he2020momentum}.

\begin{table}[t]
\centering
% \tablestyle{2.5pt}{1.1}
% \tablestyle{6.5pt}{1.1}
\unitablestyle
% \small
   \begin{tabular}{llc|cc}
   \vspace{-3pt}
  &  &  &  \multicolumn{2}{c}{OTB}\\
  Method
  & Backbone
  & Dataset
  & Precision & Success   \\
  \shline
  Supervised~\cite{he2016deep} & ResNet-18   & ImageNet & 61.4 & 43.0 \\
  SimSiam~\cite{chen2020exploring} & ResNet-18           &  ImageNet & 58.8 & 42.9 \\
  MoCo~\cite{he2020momentum}         & ResNet-18           & ImageNet & 62.0 & 47.0 \\
  VINCE~\cite{gordon2020watching}   & ResNet-18            & R2V2 & 62.9 & 46.5 \\
  Ours  & ResNet-18            & R2V2-S & $\mathbf{65.6}$ & $\mathbf{48.6}$ \\
%   \hhline{=====}
  \shline
  Supervised~\cite{he2016deep} & ResNet-50           & ImageNet & 65.8 & 45.5 \\
  SimSiam~\cite{chen2020exploring} & ResNet-50           & ImageNet & 61.0 & 43.2 \\
  MoCo~\cite{he2020momentum}         & ResNet-50     & ImageNet & 63.7 & 46.5 \\
  SeCo*~\cite{yao2020seco}        & ResNet-50 & Kinetics & 71.9 & 51.8 \\
  VINCE~\cite{gordon2020watching}   & ResNet-50      & R2V2 & 40.2 & 30.0 \\
  Ours        & ResNet-50 & R2V2 & 69.3 & 49.2 \\
  \end{tabular}
  \vspace{-1em}
\caption{Visual Object Tracking performance on OTB-100 compared with other unsupervised representation pretrain methods. SiamFC with one addtional 1x1 convolution is added. Note that concurrent work SeCo* uses two-stage training strategy, where it uses MoCo pretrained on ImageNet as first stage.}
\label{tb:tracking}
\end{table}

\subsection{Transfer to visual object tracking}
% \color{red}
We perform object tracking on the learned representations. SiamFc~\cite{bertinetto2016fully} is used as tracking method, which consists of one 1x1 convolution
upon the pre-trained frozen representations. The training is performed on GOT-10k~\cite{huang2019got} dataset and we test it on OTB2015~\cite{wu2013online}. 
The results are shown in Table~\ref{tb:tracking}. As we could see, 
when using ResNet-18 as backbone, our method outperforms previous unsupervised representation methods, as well as ImageNet supervised one, obtaining 65.6 precision, an 2.7 improvement over VINCE~\cite{gordon2020watching}, which trains on the same dataset. Our method also surpasses previous contrastive learning methods when using ResNet-50 as backbone, and achieves 3.5 precision improvement over ImageNet supervised pretrain model.
Note that we observed a performance drop when using intra-image objective alone when switching from ResNet-18 to ResNet-50. This also occurred in VINCE~\cite{gordon2020watching} which uses the same dataset for pretraining. We conjure that it is because there are only 4 frames per video, covering a long temporal window, making it hard to find correspondence within the same video, which is important for tracking task.
Although our method is slightly inferior to SeCo~\cite{yao2020seco}. However, SeCo utilizes a two-stage training strategy, where it first uses MoCo~\cite{he2020momentum} method training on ImageNet dataset, and then trains Kinetics. Our method completely trains from scratch on R2V2, which makes SeCo is not directly comparable to ours.

In order to validate the effectiveness of our method, we conduct ablation study on different loss components with respect to the tracking performance, the results are shown in Table~\ref{tb:otb_abl}. It shows that our proposed
$\mathcal{L}_{\text{cycle}}$ clearly improves methods considering only intra-image or intra-video constrative learning. Note that when training with only intra-image invariance without using temporal invariance in videos in R2V2 dataset, the performance for tracking will drop significantly, which is consistent with the results shown in VINCE~\cite{gordon2020watching}.

\begin{table}[t]
\centering
% \small
% \tablestyle{6.5pt}{1.1}
\unitablestyle
   \begin{tabular}{cccc|cc}
%   \vspace{-3pt}
  \multicolumn{3}{c}{Invariance} & & \multicolumn{2}{c}{OTB} \\
  {\tablestyle{0pt}{.9} \begin{tabular}{c} {intra} \\ {image} \end{tabular}} 
  & {\tablestyle{0pt}{.9} \begin{tabular}{c} {intra} \\ {video} \end{tabular}} 
  & {\tablestyle{0pt}{.9} \begin{tabular}{c} {cross} \\ {video} \end{tabular}} 
  & 
  Backbone
  & Precision   & Success   \\
  \shline
  \cmark & & & ResNet-18 & 53.7 & 41.2 \\
  \cmark & \cmark & & ResNet-18 & 60.1 & 45.5 \\
  \cmark & \cmark & \cmark & ResNet-18 & \textbf{65.6} & \textbf{48.6} \\
  \shline
  \cmark & & & ResNet-50 & 47.4 & 34.4 \\
  \cmark & \cmark & & ResNet-50 & 68.4 & 48.9 \\
  \cmark & \cmark & \cmark & ResNet-50 & \textbf{69.3} & \textbf{49.2} \\
  
  \end{tabular}
  \vspace{-1em}
  \caption{Ablation of different losses components (invariance target) for OTB-100 tracking on frozen features. Representations are pretrained on R2V2-S with ResNet-18, and R2V2 with ResNet-50.
  We clearly see that our proposed method with cross-video cycle consistency learning target achieves the best performance.}
  \label{tb:otb_abl}
\end{table}

\subsection{Transfer to image classification}

\subsubsection{Comparison to state of the art}
To further show the generalizability of our method, we transfer the learned representations to perform static image classification task. 
We use ImageNet dataset~\cite{deng2009imagenet} to validate our method. We apply one fully-connected layer on frozen representation as linear probing setting in~\cite{he2020momentum, gordon2020watching}.
The results are shown in Table~\ref{tb:IN}. As we could see, our method achieves 55.63\% ImageNet Top-1 accuracy, which is 1.23\% improvement over VINCE~\cite{gordon2020watching}, and 2\% improvement over MoCo~\cite{he2020momentum} under the same setting, which shows the effectiveness of our proposed method for learning image representation.

\begin{table}[t]
\centering
% \tablestyle{5pt}{1.1}
% \tablestyle{6pt}{1.1}
\unitablestyle
    \begin{tabular}{ccc|c}
    & & & ImageNet \\
    Methods & Backbone & Dataset & Top-1 (\%) \\
    \shline
    Supervised & ResNet-50 & ImageNet & 76.2 \\ 
    MoCo~\cite{he2020momentum} & ResNet-50 & ImageNet & 67.7 \\ 
    MoCo~\cite{he2020momentum} & ResNet-50 & R2V2 & 53.6 \\ 
    VINCE~\cite{gordon2020watching} & ResNet-50 & R2V2 & 54.4 \\ 
    Ours  & ResNet-50 & R2V2 & \textbf{55.6}  \\
    \end{tabular}
  \vspace{-1em}
\caption{Linear classification on frozen features results on \textbf{ImageNet} compared with other unsupervised representation learning methods.}
\label{tb:IN}
\end{table}

\subsubsection{Ablation Study}

In this section, we design various experiments to show how each of the components of our method affects performance. We use ImageNet, as well as ImageNet-100 dataset~\cite{deng2009imagenet} as our transferring dataset for ablation study, which has 126k train images and 20k test images from 100 classes. 

\paragraph{Effect of different loss components.}
In this part, we study the effect of using different losses to perform unsupervised representation learning and validate the performance on transferring to ImageNet classification.
We study three different target of invariance learning, correspond to three loss functions, (a) $\mathcal{L}_{\text{intra-image}}$ , (b) $\mathcal{L}_{\text{intra-video}}$, 
(c) $\mathcal{L}=\mathcal{L}_{\text{intra-video}}+\lambda\mathcal{L}_{\text{cycle}}$ ($\lambda=0.1$). 
The unsupervised representation training is performed on R2V2-S (ResNet-18), R2V2 (ResNet-50) dataset. The results are shown in Table~\ref{tb:in100_abl}

As we could see, using a intra-video invariance learning target $\mathcal{L}_{\text{intra-video}}$ improves the ImageNet Top-1 accuracy from 33.0\% to 33.1\%, 
indicating that frames within the same video are natural views of the query image. Furthermore, our method, which adds $\mathcal{L}_{\text{cycle}}$ to perform cross-video cycle contrastive learning, 
further boosts the Top-1 accuracy to 34.4\%, which is an absolute 1.4\% improvement compared to intra-image invariance learning. With a deeper backbone ResNet-50, a similar trend is observed as the proposed full loss with $\mathcal{L}_{\text{cycle}}$ improves 1.8\% Top-1 accuracy over intra-image objective. 
This shows that adding $\mathcal{L}_{\text{cycle}}$ makes the model 
explore cross-video relation, making features of visually similar instances across frames and videos closer.

\begin{table}[t]
\centering
% \small
% \tablestyle{6pt}{1.1}
\unitablestyle
   \begin{tabular}{cccc|c}
%   \vspace{-3pt}
  \multicolumn{3}{c}{Invariance} & & 
%   {ImageNet} \\
%   {\tablestyle{0pt}{.9} \begin{tabular}{c} {ImageNet} \\ {Top-1} \end{tabular}}  \\
\\
  {\tablestyle{0pt}{.9} \begin{tabular}{c} {intra} \\ {image} \end{tabular}} 
  & {\tablestyle{0pt}{.9} \begin{tabular}{c} {intra} \\ {video} \end{tabular}} 
  & {\tablestyle{0pt}{.9} \begin{tabular}{c} {cross} \\ {video} \end{tabular}} 
  & 
  Backbone
  & 
  {\tablestyle{0pt}{.9} \begin{tabular}{c} {ImageNet} \\ {Top-1 (\%)} \end{tabular}}  \\
%   Top-1 (\%)
%   \\
  \shline
  \cmark & & & ResNet-18 & 33.0\\
  \cmark & \cmark & & ResNet-18 &33.1  \\
  \cmark & \cmark & \cmark & ResNet-18 & \textbf{34.4} \\
  \shline
  \cmark & & & ResNet-50 & 53.8 \\
  \cmark & \cmark & & ResNet-50 & 55.1 \\
  \cmark & \cmark & \cmark & ResNet-50 & \textbf{55.6}\\
  
  \end{tabular}
  \vspace{-1em}
\caption{Ablation of different losses components ImageNet classification on frozen features. Representations are pretrained on R2V2-S with ResNet-18 and R2V2 with ResNet-50. 
We clearly see that our proposed method with cross-video cycle consistency learning target achieves the best performance.}
  \label{tb:in100_abl}
\end{table}

\paragraph{Intra-image or Intra-video invariance are essential for good representation.}
While exploring cross-image or cross-video information are helping learning better representation, it should be built on representation that is 
invariant of instance or videos. We conduct experiments that use $\mathcal{L}_{\text{cycle}}$ only to learn the representation. A random view (data augmentation)
$k_{++}$ of the query $q_i$ in the neighbor set $U$. The results are shown in Table~\ref{tb:include_self}. As we could see, when the neighbor size is small (\eg 256), using 
$\mathcal{L}_{\text{cycle}}$ alone could learn representation that has better performance when transferring to linear classification on ImageNet-100, compared 
to using $\mathcal{L}_{\text{intra-video}}$. However, when the neighbor set is
large, the performance drops largely. On the other hand, if $k_{++}$ is not included in the neighbor set $U$, using a neighbor size of 256, 
and $\mathcal{L}_{\text{cycle}}$ alone would lead to 45.78\% Top-1 accuracy on ImageNet-100, which is much worse. We safely draw that when neighbor size is small and a random view $k_{++}$ of the query is included 
in the neighbor set, $\mathcal{L}_{\text{cycle}}$ \emph{degenerates} to $\mathcal{L}_{\text{intra-video}}$ by learning to make the similarity of $q_i$ and $k_{++}$ maximized.
Thus, for making the model learn truly cross-video relation, \textbf{excluding} $k_{++}$ from the neighbor set $U$ is necessary. 
However, directly learning correspondence cross videos would be hard and shows worse results of 45.78\% Top-1 accuracy on ImageNet-100. 
It is essential to add $\mathcal{L}_{\text{intra-video}}$ with $\mathcal{L}_{\text{cycle}}$ (i.e. using the full loss $\mathcal{L}$ in Eq.), boosting to 58.50 \% Top-1 accuracy on ImageNet-100.

\begin{table}[t]
\centering
% \tablestyle{6pt}{1.1}
\unitablestyle
\begin{tabular}{c|cccc}
Size of neighbor set $U$ & 128 & 256 & 512 & 16384 \\ \shline
Acc Top-1 (\%)  & 56.56 & \textbf{57.00} & 56.40 & 52.96 \\ 
Acc Top-5 (\%)  & \textbf{83.10} & 83.06 & 82.64 & 79.46  \\ 
\end{tabular}
  \vspace{-1em}
\caption{Ablation study. Accuracy on \textbf{ImageNet-100} under linear classification protocol varying the size of the neighbor set $U$. 
Representations are pretrained \textbf{using $\mathcal{L}_{\text{cycle}}$ only} on R2V2-S with ResNet-18. One view of the current query image is \textbf{included} in the neighbor set.}
\label{tb:include_self}
\end{table}

\paragraph{How to choose neighbors?}
We conduct experiments to study how the size of neighbor set $U$ is affecting the performance. Full loss $\mathcal{L}$ is used to perform unsupervised representation learning on R2V2-S dataset
 and the size of $U$ varies from 256 to 16384.
Views of the current query are not added to the neighbor set. Then we evaluate the learned representation by conducting linear classification on froze representation on ImageNet-100. 
The results are shown in Table~\ref{tb:neighbor_size_full}. As we could see, increasing the neighbor size give us better representation on behalf of better ImageNet-100 Top-1 accuracy. This is expected as 
a large neighbor size of $U$ would gives higher probability that the query could find correspondence across videos. The best Top-1 accuracy of 58.48\% is achieved when using 16384 as neighbor size and 
we use this as default setting hereafter.

\begin{table}[t]
\centering
% \tablestyle{6pt}{1.1}
\unitablestyle
\begin{tabular}{c|cccc}
Size of neighbor set $U$ & 256   & 1024  & 4096  & 16384 \\ \shline 
Acc Top-1 (\%) & 56.68 & 57.36 & 56.82 & \textbf{58.48} \\ 
Acc Top-5 (\%) & \textbf{84.04} & 83.78 & 83.52 & 83.50  \\
\end{tabular}
  \vspace{-1em}
\caption{Ablation study. Accuracy on \textbf{ImageNet-100} under linear classification protocol varying the size of the neighbor set $U$. 
Representations are pretrained \textbf{using full loss $\mathcal{L}$} on R2V2-S with ResNet-18.
}
\label{tb:neighbor_size_full}
\end{table}

We also study that if using top-K nearest neighbors to perform the reconstruction in Eq.~\ref{eq:cycle_forward} would help representation learning. Top-K neighbors are selected by 
ranking the similarity $a_{q_i, u}$ in the initial neighbor set, and the top K neighbors are chosen to construct the new neighbor set. Then the cross-video cycle consistency learning is perform using the new neighbor set. 
Then we transfer the learned representations to the task of linear classification on ImageNet-100. 
The results are shown in Table~\ref{tb:abl_top_k}. We could see that the top-1 accuracy is robust of a wide range of K, from 8 to 128. However, using top-K neighbors show worse performance compared 
to randomly chosen a neighbor set $U$ of size 16384, which has 58.50\% Top-1 accuracy on ImageNet-100. We conjure that a large and random neighbor set could have higher probability of finding visual similar images and 
could correct the model's false belief at early stage that misses useful neighbors in the top-K neighbors. Thus we randomly choose the neighbors $U$ and set the size of $U$ to 16384 as default.

\begin{table}[t]
    % \small
\centering
% \tablestyle{6pt}{1.1}
\unitablestyle
\begin{tabular}{c|ccccc}
 K & 8 & 16 & 32 & 128 & 256 \\ \shline
Acc Top-1 (\%) & 57.52 & 57.40 & \textbf{57.64}  & 57.62   &  56.98       \\ 
Acc Top-5 (\%) & 83.02 & 82.84  & 83.62  &  83.76   & \textbf{83.82}        \\ 
\end{tabular}
  \vspace{-1em}
\caption{Ablation study. Accuracy on \textbf{ImageNet-100} under linear classification protocol. Ablation on K when using top-K nearest neighbors to perform cross video cycle consistency learning. 
Representations are pretrained on R2V2-S with ResNet-18. 
}
\label{tb:abl_top_k}
\end{table}

\paragraph{Balance between intra-video invariance and cross-video relation learning.} 
We study the influence of the balancing factor $\lambda$ in the loss term $\mathcal{L}$.
We conduct unsupervised representation learning using different balancing factor $\lambda$ in Eq.~\ref{eq:loss} on R2V2-S dataset, and evaluate the performance using the task of linear classification on frozen learned representation 
on ImageNet-100 dataset. The results are shown in Table~\ref{tb:balance_weight}. We could see that while adding $\mathcal{L}_{\text{cycle}}$ helps learn better representation 
for the task of linear classification on ImageNet-100, using a relative small (\eg 0.1) is best. This also coincides our previous finding that intra-video invariance learning $\mathcal{L}_{\text{intra-video}}$
is essential for the cross-video cycle contrast loss $\mathcal{L}_{\text{cycle}}$ to build on. We set $\lambda = 0.1$ as default hereafter.

\begin{table}[t]
\centering
% \tablestyle{6pt}{1.1}
\unitablestyle
\scalebox{1.00}{
\begin{tabular}{c|cccccc}
$\lambda$ & 0.05  & 0.1   & 0.3   & 0.5   & 0.7 & 1.0 \\ \shline
Acc Top-1 (\%) & 57.16 & \textbf{58.48} & 57.04 & 57.90 & 57.48    & 57.84    \\ 
Acc Top-5 (\%)& 83.08 & 83.50 & 83.80 & 83.50  & 83.18    &  \textbf{83.70}   \\
\end{tabular}
}
\caption{Ablation results on loss balancing factor $\lambda$ in Eq.~\ref{eq:loss}, Top-1 and Top-5 accuracy results of \textbf{ImageNet-100} linear classification on frozen representations are shown.
Representations are pretrained on R2V2-S with ResNet-18. 
}
\label{tb:balance_weight}
\end{table}

\subsection{Transfer to video action recognition}
% \vspace{-1em}

\subsubsection{Comparison to state of the art}
We evaluate the learned feature representation on the task of video action recognition on UCF101~\cite{soomro2012ucf101} dataset. UCF101 has 13320 videos from 101 action categories. We train and test our model on split1 of UCF101.
For simplicity, we directly use ResNet other than 3D convolution based methods. The video representation is obtained by averaging the frame representation from the video, and one  
fully-connected layer is used for predicting the action class on this video representation. Following~\cite{morgado_avid_cma}, multiple clips are sampled from the video and the predictions of the clips are averaged for the final results.  The results are shown in Table~\ref{tb:UCF101}. As we could see, our method is able to achieve 76.8\% top-1 accuracy on UCF101 with ResNet-18 as backbone, surpassing the previous best one of 68.2\% with similar number of parameters.

We also list some results of methods with large 3D ConvNets models in Table~\ref{tb:UCF101}. Our method is not directly comparable to these methods as we use 2D ConvNets, which not only have less parameters but also much less FLOPs compared to 3D ConvNets. We emphasize that our representation is able to solve multiple downstream tasks while the previous 3D ConvNets are designed only for action recognition. Note that, while DPC with large 3D-ResNet34 (32.6M) as backbone achieves 75.7\%, our model with fewer parameters (11.7M), is able to surpass DPC with 3D-ResNet18 (14.2M) as backbone. Notably, when using ResNet-50 as backbone, our method is able to achieve 82.1\%, surpassing other methods with larger 3D ConvNets models (\eg MemDPC) as well. 
The results validate 
the effectiveness and transferability of the learned representations of our method.

\begin{table*}[t]
\centering
% \small
% \tablestyle{6pt}{1.1}
\unitablestyle
\begin{tabular}{llll|c}
Method      & Backbone  &\#Param  & Dataset     & UCF101    \\ 
\shline

Random Initialization~\cite{han2019video}           & 3D-ResNet18 &14.2M    & -               & 46.5                      \\ 
ImageNet Supervised Pretrained~\cite{simonyan2014two}        & VGG-M-2048 & 353M  & -               & 73.0                    \\  \hline

Shuffle~\&~Learn~\cite{misra2016shuffle}~($227 \times 227$) & AlexNet &61M  & UCF101/HMDB51   & 50.2                       \\
OPN~\cite{lee2017unsupervised}  & AlexNet  & 61M & UCF101 & 56.3                       \\
DPC~\cite{han2019video}($128 \times 128$)                     & 3D-ResNet18 &14.2M & UCF101    & 60.6     \\ 
\hline
3D-RotNet~\cite{jing2018self}~($112 \times 112$)         & 3D-ResNet18-full &33.6M & Kinetics-400    & 62.9                    \\
3D-ST-Puzzle~\cite{kim2019self}~($224 \times 224$)     & 3D-ResNet18-full&33.6M  & Kinetics-400    & 63.9 \\
SpeedNet~\cite{benaim2020speednet}~($224 \times 224$) & I3D&12.1M & Kinetics-400 & 66.7 \\
DPC~\cite{han2019video}($128 \times 128$)                   & 3D-ResNet18 &14.2M & Kinetics-400    & 68.2   \\ 

DPC~\cite{han2019video}($224 \times 224$)                   & 3D-ResNet34 & -  & Kinetics-400    & 75.7     \\ 
Video-Pace~\cite{wang2020self} & R(2+1)D &33.3M & Kinetics-400 & 77.1\\
CBT~\cite{sun2019learning} (112x112) &S3D & - & Kinetics-400 & 79.5 \\
MemDPC~\cite{han2020memory}  &  R-2D3D & 32.4M & Kinetics-400 & 78.1 \\

% VTHCL~\cite{yang2020video} & 3D-ResNet18 &14.2M & Kinetics-400 & 80.6 \\

Temporal-ssl~\cite{jenni2020video} & R(2+1)D &33.3M & Kinetics-400 & 81.6\\
VTHCL~\cite{yang2020video} & 3D-ResNet50 & 31.7M & Kinetics-400 & 82.1 \\

\hline

Ours($224 \times 224$)  & ResNet-18 &$\mathbf{11.69}$ M& R2V2 & $76.8$ \\
Ours($224 \times 224$)  & ResNet-50 &$\mathbf{25.56}$ M& R2V2 & $\mathbf{82.1}$ \\

\shline
\rowstyle{\color{gray}}
XDC~\cite{alwassel2019self} (224x224) & R(2+1)D &33.3M & Kinetics-400 & 86.8 \\
AVID+CMA~\cite{morgado_avid_cma}(224x224) & R(2+1)D & 33.3M & Kinetics-400 & 87.5 \\ 
CVRL~\cite{qian2020spatiotemporal} & 3D-ResNet50 & 31.7M & Kinetics-400 & 92.9 \\
$\rho$ BYOL~\cite{feichtenhofer2021large} & 3D-ResNet50 & 31.8M & Kinetics-400 &95.5 \\
% CoCLR-RGB~\cite{Han20}& S3D-23 & & K400 & 87.9 \\
% VTHCL~\cite{yang2020video}& 3D-ResNet18 & 14.2M & Kinetics-400& 82.1 \\
% 		CVRL [Qian et al. 2020] & 3D-ResNet50 & K400 & \textbf{92.9}\\

\end{tabular}
  \vspace{-1em}
\caption{Video action recognition accuracy comparison with other unsupervised representation methods on \textbf{UCF101}. We compare with methods using RGB modality. We mainly compare with models that has \textbf{similar parameters} as ours and list some large 3D ConvNet models (in gray) for reference.}
\label{tb:UCF101}
\end{table*}

\subsubsection{Ablation Study}
% {Ablation of losses on video action recognition on UCF101.}
We study the effect of different losses on the task of video action recognition on UCF-100 dataset. The pre-trained representations are fixed and a single fully-connected layer is added to predict the 
action class upon the averaged frame representations.
The results are shown in Table~\ref{tb:ucf_abl}. As we could see, for video action recognition task, it is beneficial to learn representations that are invariant within the same video, compared to 
invariant to the same image, rises the Top-1 accuracy from 45.1\% to 48.4\%, with ResNet-18 as backbone. However, it is also beneficial to make representations of similar videos closer other than separating them, as adding our cross-video cycle-contrastive 
learning loss $\mathcal{L}_{\text{cycle}}$ further improves the Top-1 accuracy to 50.5\%. Similarly, when using ResNet-50 as backbone, adding our $\mathcal{L}_{\text{cycle}}$ achieves the best results of 67.1\%. 
The results validate that our proposed cross-video cycle-contrastive learning target can learn representations that 
transfer well to video recognition tasks.

\begin{table}[t]
\centering
% \small
% \tablestyle{6pt}{1.1}
\unitablestyle
   \begin{tabular}{cccc|c}
%   \vspace{-3pt}
  \multicolumn{3}{c}{Invariance} & & 
%   {ImageNet} \\
%   {\tablestyle{0pt}{.9} \begin{tabular}{c} {ImageNet} \\ {Top-1} \end{tabular}}  \\
\\
  {\tablestyle{0pt}{.9} \begin{tabular}{c} {intra} \\ {image} \end{tabular}} 
  & {\tablestyle{0pt}{.9} \begin{tabular}{c} {intra} \\ {video} \end{tabular}} 
  & {\tablestyle{0pt}{.9} \begin{tabular}{c} {cross} \\ {video} \end{tabular}} 
  & 
  Backbone
  & 
  {\tablestyle{0pt}{.9} \begin{tabular}{c} {UCF101} \\ {Top-1 (\%)} \end{tabular}}  \\
%   Top-1 (\%)
%   \\
  \shline
  \cmark & & & ResNet-18 & 45.1\\
  \cmark & \cmark & & ResNet-18 & 48.4 \\
  \cmark & \cmark & \cmark & ResNet-18 & \textbf{50.5} \\
  \shline
  \cmark & & & ResNet-50 &  62.2\\
  \cmark & \cmark & & ResNet-50 &66.2  \\
  \cmark & \cmark & \cmark & ResNet-50 & \textbf{67.1}\\
  
  \end{tabular}
  \vspace{-1em}
    \caption{Ablation of different losses components UCF action recognition learned on frozen features. Representations are pretrained on R2V2 with ResNet-18 and ResNet-50. 
    }
  \label{tb:ucf_abl}
\end{table}

\subsubsection{Nearest Neighbor evaluation}

To further validate our representation could learn cross-video information, we perform nearest neighbor retrieval experiments on the learned representation on UCF101 dataset on both frame level and clip level.

For frame retrieval experiments, following~\cite{buchler2018improving}, 10 frames are sampled for each video. The representations of the frames from test set are used to find the nearest neighbors on the train set. 
For clip retrieval experiments, following~\cite{xu2019self}, 10 clips per video are sampled. The clips extracted from the test set are used to find nearest neighbors on the train set. Cosine distance of representations is used as ranking 
criterion. If the class of the query sample appears in the class set of the k nearest neighbors, it is considered as a correct retrieval.

The results are shown in Table~\ref{tb:ucf101_clip_retrieval}. Our model is able to surpass previous methods largely on both
frame retrieval and clip retrieval experiments. Notably, our method with ResNet-18 as backbone has a superior accuracy when considering small k (\eg k=1, 5, 10). For instance, our method achieves a top-1 accuracy of 45.8\% on frame retrieval experiments, 
where previous best result from Buchler \etal~\cite{buchler2018improving} is 25.7\%, which is an absolute 20.1\% improvement. For clip retrieval, our method has a top-1 accuracy of 39.7\%, an absolute 25.6\% improvement compared to previous 
best of 14.1\%. This indicates that our method is able to make the representations of similar frames/videos (short clips) closer, as they have higher probability belonging to the same class.
Furthermore, our model is trained solely on R2V2 dataset and has never seen samples from UCF101 dataset, while other methods is trained on UCF101.
The results indicate the transferability of our method. Video representations are obtained by averaging clip representations. We could see that even though our model does not
have ground truth class to guide the representation learning, it manages to make video representations from the same class or having similar visual structures close.

\begin{table*}[t]
\centering
% \tablestyle{6pt}{1.1}
\unitablestyle
\begin{tabular}{lc|ccccc}
Methods & Training Data & Top-1 & Top-5 & Top-10 & Top-20 & Top-50 \\ 
\shline
\multicolumn{2}{l|}{\emph{Frame retrieval}:}\\
\midrule
Jigsaw~\cite{noroozi2016unsupervised}(CaffeNet)  & ImageNet & 19.7 & 28.5 & 33.5 & 40.0 & 49.4 \\
OPN~\cite{lee2017unsupervised}(CaffeNet)  & UCF101 & 19.9 & 28.7 & 34.0 & 40.6 & 51.6 \\
Buchler~\cite{buchler2018improving}(CaffeNet) & ImageNet+UCF101& 25.7 & 36.2 & 42.2 & 49.2 & 59.5 \\ 
% Ours (ResNet-18) & R2V2-S & 41.5 & 51.8 & 56.7 & 62.3 & 71.0 \\ 
Ours (ResNet-18) & R2V2 & 45.8 & 56.2 & 61.4 & 67.0 & 75.2 \\ 
Ours (ResNet-50) & R2V2 & \textbf{52.6} & \textbf{63.3}&\textbf{68.1} & \textbf{73.3}  & \textbf{80.6} \\ 
% \midrule
\hhline{=======}
    \multicolumn{2}{l|}{\emph{Clip retrieval}:}\\
    \midrule
% ResNet18 (random) & & 14.3  & 22.8 & 27.9 & 34.2 & 43.6 \\
Order~\cite{xu2019self} (C3D) & UCF101 & 12.5 & 29.0 &39.0 &50.6 &66.9 \\
Order~\cite{xu2019self} (R(2+1)D) & UCF101 & 10.7 &25.9 &35.4& 47.3& 63.9\\
Order~\cite{xu2019self} (R3D) &  UCF101 & 14.1 & 30.3 &40.0 &51.1 & 66.5 \\
SpeedNet~\cite{benaim2020speednet} (S3D-G) & Kinetics-400 & 13.0 & 28.1 & 37.5 & 49.5&  65.0 \\
% Ours (ResNet-18) & R2V2-S & 35.4 & 44.7 & 50.0 & 56.1 & 65.5 \\
Ours (ResNet-18) & R2V2 & 39.7 & 50.3 & 55.9 & 62.0 & 70.7 \\
Ours (ResNet-50) & R2V2 & \textbf{46.8} & \textbf{56.7} & \textbf{62.1} & \textbf{67.6} & \textbf{75.1} \\
\end{tabular}
% 	\vspace{1em}
  \vspace{-1em}
\caption{Frame retrieval and Clip retrieval results on \textbf{UCF101} compared to other unsupervised representation learning methods. Results of Jigsaw and OPN are taken from~\cite{buchler2018improving}.
Our model largely surpasses previous methods and manages to do so without using UCF101 samples.}
\label{tb:ucf101_clip_retrieval}
\end{table*}

Overall, we design various ablation experiments to study different components of the loss, including image recognition, video recognition, video retrieval, tracking. 
The results validate the effectiveness of our proposed cross-video cycle-contrastive learning target loss $\mathcal{L}_{\text{cycle}}$.

% \vspace{-0.05}
\section{Conclusion}
% \vspace{-0.05}

In this paper, we propose a cross-video cycle-consistent contrastive learning objective to perform self-supervised learning for image representations. 
The proposed method could learn to explore cross-video relations to make not only the representations of images within the same video closer, but also
representations from different videos with similar visual structures close, without using ground truth class labels or generating  pseudo labels. We perform self-supervised representation learning on unlabeled R2V2 video dataset, and show that the learned image representation transfer well to multiple downstream tasks including visual tracking, image classification and action recognition. Extensive ablation studies are conducted and validate the effectiveness of our proposed method. We hope our approach can open up an  opportunity to utilize cross-instance paired data for learning general image representations.

{\small
\bibliographystyle{ieee_fullname}
\bibliography{egbib}
}

\end{document}